\documentclass[11pt,a4paper]{article}
\usepackage[hyperref]{emnlp2018}
\usepackage{times}
\usepackage{graphicx}
\usepackage{latexsym}
\usepackage{graphicx}

\usepackage{url}

\aclfinalcopy 

\usepackage{tikz}
\newcommand*\circled[1]{\tikz[baseline=(char.base)]{
            \node[shape=circle,draw,inner sep=2pt] (char) {#1};}}

\title{Demonstrating \textsc{Par4Sem} -  \\ A Semantic Writing Aid with Adaptive Paraphrasing   }

\author{Seid Muhie Yimam \qquad Chris Biemann \\
Language Technology Group\\ 
Department of Informatics\\ 
Universit\"{a}t Hamburg, Germany\\
 {\tt \{yimam,biemann\}@informatik.uni-hamburg.de}
}

\date{}

\begin{document}
\maketitle
\begin{abstract}
  In this paper, we present \textsc{Par4Sem}, a semantic writing aid tool based on adaptive paraphrasing. Unlike many annotation tools that are primarily used to collect training examples, \textsc{Par4Sem} is integrated into a real word application, in this case a writing aid tool, in order to collect training examples from usage data. \textsc{Par4Sem} is a tool, which supports an adaptive, iterative, and interactive process where the underlying machine learning models are updated for each iteration using new training examples from usage data. After motivating the use of ever-learning tools in NLP applications, we evaluate \textsc{Par4Sem} by adopting it to a text simplification task through mere usage. 
\end{abstract}

\section{Introduction}
Natural language processing and semantic applications that depend on a machine learning component require training data, i.e. examples from which the machine learning algorithm learns from. The training datasets require, most of the time, manual annotation. Usually, such annotations are conducted in a predefined cycle of annotation activities. Once the annotation problem is identified, a standalone annotation tool along with the annotation guideline is developed. At the end of the annotation cycle, the collected dataset is fed to the machine learning component, which produces a static model that can be used thereafter in an application.

 Possible limitations of these annotation approaches are:  1) Developing a standalone annotation tool is costly, sometimes expert or specially trained annotators are required. 2) There is no direct way to evaluate the dataset towards its effectiveness for the real-world application. 3) It suffers from what is known as \emph{concept drift}, as the annotation process is detached from the target application, the dataset might get obsolete over time. 

In this regard, we have dealt specifically with the semantic annotation problem, using an adaptive, integrated, and personalized annotation process. By \emph{adaptive}, we mean that target applications do not require pre-existing training data, rather it depends on the usage data from the user. The machine learning model then adapts towards the actual goal of the application over time. Instead of developing a standalone annotation tool, the collection of training examples is \emph{integrated} inside a real-world application.  Furthermore, our approach is \emph{personalized} in a sense that the training examples being collected are directly related to the need of the user for the application at hand.  After all, the question is not: how good is the system \emph{today}? It is rather: how good will it be \emph{tomorrow} after we use it today?

Thus, such adaptive approaches have the following benefits:

	\noindent \textbf{Suggestion and correction options}: Since the model immediately starts learning from the usage data, it can start predicting and suggesting recommendations to the user immediately. Users can evaluate and correct suggestions that in turn help the model to learn from these corrections. \\
	\noindent \textbf{Less costly}: As the collection of the training data is based on usage data, it does not need a separate annotation cycle.  \\
	\noindent \textbf{Personalized}: It exactly fits the need of the target application, based on the requirement of the user. \\
	\noindent \textbf{Model-Life-Long Learning}: As opposed to static models that once deployed on the basis of training data, adaptive models incorporate more training data the longer they are used, which should lead to better performance over time.

We have developed \textsc{Par4Sem}, a semantic writing aid tool using an adaptive paraphrasing component, which is used to provide context-aware lexical paraphrases while composing texts. The tool incorporates two adaptive models, namely target identification and candidate ranking. The adaptive target identification component is a classification algorithm, which learns how to automatically identify target units (such as words, phrases or multi-word expressions), that need to be paraphrased. When the user highlights target words (\emph{usage data}), it is considered as a training example for the adaptive model.

The adaptive ranking model is a learning-to-rank machine learning algorithm, which is used to re-rank candidate suggestions provided for the target unit. We rely on existing paraphrase resources such as PPDB 2.0, WordNet, distributional thesaurus and word embeddings (see Section \ref{parresource}) to generate candidate suggestions. 

Some other examples for adaptive NLP setups include: 1) online learning for ranking, example \newcite{Yogatama2014} who tackle the pairwise learning-to-ranking problem via a scalable online learning approach, 2) adaptive machine translation (MT), e.g. \newcite{W14-0311} describe a framework for building adaptive MT systems that learn from post-editor feedback, and 3) incremental learning for spam filtering, e.g. \newcite{Sheu2017} use a window-based technique to estimate for the condition of concept drift for each incoming new email.

We have evaluated our approach with a lexical simplification task use-case. The lexical simplification task contains complex word identification (\emph{adaptive target identification}) and simpler candidate selection (\emph{adaptive ranking}) components. 

As far as our knowledge concerned, \textsc{Par4Sem} is the first tool in the semantic and NLP research community, where adaptive technologies are integrated into a real-world application. \textsc{Par4Sem} is open source\footnote{\url{https://uhh-lt.github.io/par4sem/}} and the associated data collected for the lexical simplification use-case are publicly available. The live demo of \textsc{Par4Sem} is available at \url{https://ltmaggie.informatik.uni-hamburg.de/par4sem}.

\section{System Architecture of \textsc{Par4Sem}}
\begin{figure}[ht]
		\includegraphics[width=1.0\linewidth]{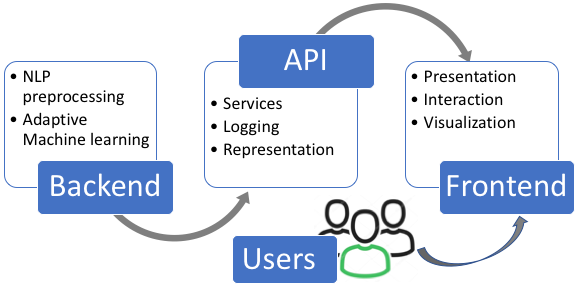}
		\caption{The main components of \textsc{Par4Sem}  }
		\label{fig:par4semgen} 
\end{figure}

The \textsc{Par4Sem} system  consists of backend, frontend, and API components. The backend component is responsible for NLP related pre-processing, adaptive machine learning model generation, data storage, etc. The frontend component sends requests to the backend, highlights target units, presents candidate suggestions, sends user interaction to the database and so on. The API component transforms the frontend requests to the backend and returns responses to the frontend. Figure \ref{fig:par4semgen} shows the three main components of \textsc{Par4Sem} and their interactions.

\subsection{The Backend Component}
The backend component consists of several modules. For the adaptive paraphrasing system, the first component is to identify possible target units (such as single words, phrases, or multi-word expressions). For our lexical simplification use-case, the target units identification component is instantiated with the datasets obtained from \newcite{Yimam:2017:ijcnlp,R17-1104,W18-0507}. The adaptive target identification unit then keeps on learning from the usage data (when the user highlights portions of the text to get paraphrase candidate suggestions).

Once target units are marked or recognized (by the target unit identification system), the next step is to generate possible candidate suggestion for the target unit (paraphrase candidates). The candidate suggestion module includes candidate generation and candidate ranking sub-modules. Section \ref{parresource} discusses our approaches to generating and ranking paraphrase candidates in detail.  

\subsubsection{Paraphrasing Resources}
\label{parresource}
Paraphrase resources are datasets where target units are associated with a list of candidate units equivalent in meaning, possibly ranked by their meaning similarity. One can refer to the work of \newcite{Ho2014} about the details on how paraphrase resources are produced, but we will briefly discuss the different types of paraphrase resources that are used in generating candidate suggestions for \textsc{Par4Sem}.

	\noindent  \textbf{PPDB 2.0}: The Paraphrase Database (PPDB) is a collection of over 100 million paraphrases that was automatically constructed using a bilingual pivoting method.  Recently released PPDB 2.0 includes improved paraphrase rankings, entailment relations, style information, and distributional similarity measures for each paraphrase rule \cite{P15-2070}. \\
	\noindent  \textbf{WordNet}: We use WordNet synonyms, which are described as \emph{words that denote the same concept and are interchangeable in many contexts} \cite{Miller:1995:WLD:219717.219748}, to produce candidate suggestions for a given target unit. \\
	\noindent  \textbf{Distributional Thesaurus -- JoBimText}: We use JoBimText, an open source platform for large-scale distributional semantics based on graph representations \cite{TUD-CS-2013-0338}, to extract candidate suggestions that are semantically similar to the target unit. \\
	\noindent  \textbf{Phrase2Vec}: We train a Phrase2Vec model \cite{mikolov2013distributed} using English Wikipedia and the AQUAINT corpus of English news text \cite{david2002}. \newcite{mikolov2013distributed} pointed out that it is possible to extend the word based embeddings model to phrase-based model using a data-driven approach where each phrase or multi-word expressions are considered as individual tokens during the training process. We have used a total of 79,349 multiword expression and phrase resources as given in \newcite{yimam-EtAl:2016:MWE}. We train the Phrase2Vec embeddings with 200 dimensions using skip-gram training and a window size of 5. We have retrieved the top 10 similar words to the target units as candidate suggestions.
	 
\subsubsection{Adaptive Machine Learning}
\label{machinelearning}
\textsc{Par4Sem} incorporates two adaptive machine learning models. The first one is used to identify target units (\emph{target adaption}) in a text while the second one is used to rank candidate suggestions (\emph{ranking adaption}). Both models make use of usage data as a training example. The target adaption model predicts target units based on the usage data (training examples) and sends them to the frontend component, which are then highlighted for the user. If the user replaced the highlighted target units, they are considered as positive training examples for the next iteration.

\begin{figure}[ht]
		\includegraphics[width=1.0\linewidth]{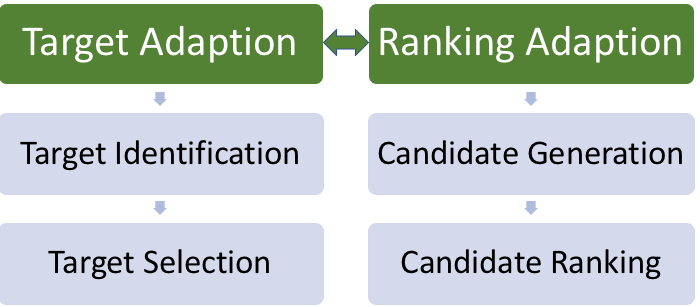}
		\caption{The main and sub-processes of target and ranking adaption components of \textsc{Par4Sem}. }
		\label{fig:par4semmodels} 
\end{figure}

\begin{figure}[ht]
		\includegraphics[width=1.0\linewidth]{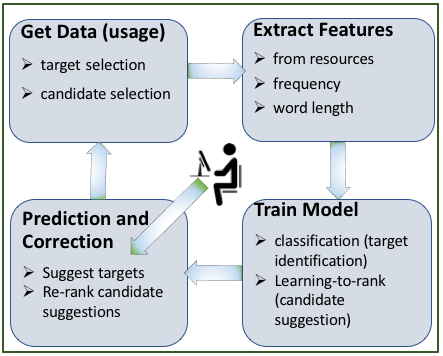}
		\caption{The loop for the generation of the adaptive models of \textsc{Par4Sem}.  }
		\label{fig:par4sempipe} 
\end{figure}

The ranking adaption model first generates candidate paraphrases using the paraphrase resource datasets (see Section \ref{parresource}). As all the candidates generated from the paraphrase resources might not be relevant to the target unit at a context, or as the number of candidates to be displayed might be excessively large (for example the PPDB 2.0 resource alone might produce hundreds of candidates for a target unit), we re-rank the candidate suggestions using a learning-to-rank adaptive machine learning model. 
Figure \ref{fig:par4semmodels} displays the process of the adaptive models while Figure \ref{fig:par4sempipe} displays the pipeline (as a loop) used in the generations of the adaptive models.

\begin{figure}[ht]
		\includegraphics[width=1.0\linewidth]{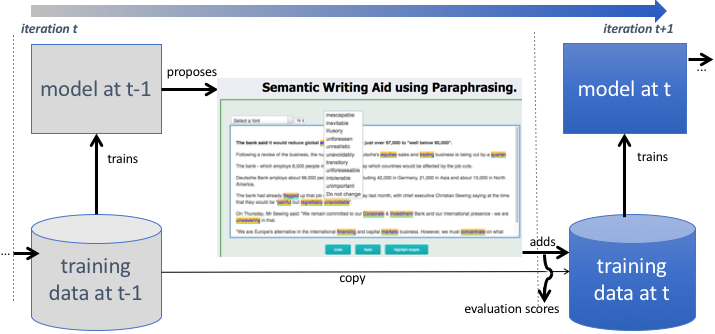}
		\caption{The iterative and adaptive interaction of \textsc{Par4Sem}.  }
		\label{fig:par4semadapt} 
\end{figure}

The whole process is iterative, interactive, and adaptive in a sense that the underlying models (both target adaption and ranking adaption) get usage data continuously from the user. The models get updated for each iteration, where $n$ examples conducted in a batch mode without model update, and provide better suggestions (as target units or candidate suggestions) for the next iteration. The user interacts with the tool, probably accepting or rejecting tool suggestions, which is fed as a training signal for the next iteration’s model. Figure \ref{fig:par4semadapt} shows the entirety of interactions, iterations, and adaptive processes of the \textsc{Par4Sem} system. In the first iteration, the ranking is provided using a baseline language model while for the subsequent iterations, the usage data from the previous batches (\emph{$t$-$1$}) is used to train a model that is used to rank the current batch ($t$). 

\subsubsection{Backend Technologies}
The backend components are fully implemented using the Java programming language. Text segmentation such as sentence splitting, tokenization, lemmatization, and parts of speech tagging is handled using the Apache OpenNLP\footnote{\url{https://opennlp.apache.org/}} library. 

For the target unit identification system, we have used \emph{Datumbox}\footnote{\url{http://www.datumbox.com/}}, a powerful open-source machine learning framework written in Java. We have used specifically the \emph{Adaboost} classification algorithm.  

For the ranking model, RankLib, which is the well-known library for the learning to rank algorithms from the Lemur\footnote{\url{https://sourceforge.net/p/lemur/wiki/RankLib/}} project is integrated. All the data related to \textsc{Par4Sem} interactions (usage data, time, and user details) are stored in a MySQL database.
\subsection{Frontend Components}
The frontend component of \textsc{Par4Sem} is designed where document composing with a semantic paraphrasing capability is integrated seamlessly. It is a web-based application allowing access either on a local installation or over the internet.     

\subsubsection{UI Components for Paraphrasing}
\begin{figure}[ht]
		\includegraphics[width=1.0\linewidth]{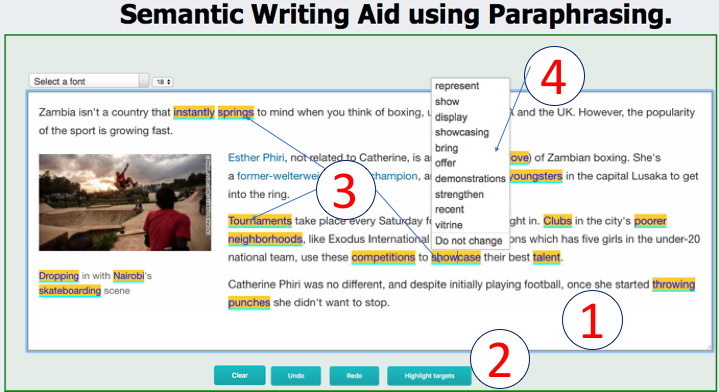}
		\caption{The \textsc{Par4Sem} text editing component that is used to compose texts, highlight target units, and display candidate suggestions for the target units.}
		\label{fig:par4semedit} 
\end{figure}

The frontend part of \textsc{Par4Sem} comprises different modules. The most important UI component is the text editing interface (Figure \ref{fig:par4semedit}) that allows for adding text, highlighting target units, and displaying candidate suggestions.  \circled{{\color{red}1}} is the main area to compose (or paste) texts. The operational buttons (\circled{{\color{red}2}}) are used to perform some actions such as  to undo and redo (composing, target unit highlighting, and paraphrase ranking), automatically highlighting target units, and clear the text area. Target units are underlined in cyan color and highlighted in yellow background color as a link (\circled{{\color{red}3}}) which enables users to click, display, and select candidate suggestions for a replacement (\circled{{\color{red}4}}).

\subsubsection{Frontend Technologies}
The frontend components are implemented using HTML, CSS and JavaScript technologies. For the text highlighting and candidate suggestion replacement, the jQuery Spellchecker\footnote{\url{http://jquery-spellchecker.badsyntax.co/}} module is slightly modified to incorporate the semantic highlighting (underline in cyan and a yellow background). The accompanied documentation and datasets of \textsc{Par4Sem}\footnote{\url{https://uhh-lt.github.io/par4sem/}} are hosted at Github pages.
\subsection{RESTful API Component}
Semantic technologies, those like \textsc{Par4Sem} incorporates highly dynamic dimensions. One dimension is that the paraphrase resources can be varied depending on the need of the application. Another dimension is that the application can be in different languages. If the backend and the frontend technologies are highly coupled, it will be difficult to reuse the application for different languages, resources, and applications. To solve this problem, we have developed \textsc{Par4Sem} using a RESTful API (aka. microservices) as a middleware between the backend and the frontend components.

The API component consumes requests (getting target units and candidate suggestions) or resources (saving usage data such as selection of new target units, user's preference for candidate ranking, user and machine information) from the frontend and transfers them to the backend. The backend component translates the requests or resources and handles them accordingly. Spring Boot\footnote{\url{https://projects.spring.io/spring-boot/}} is used to implement the API services.
\subsubsection{Installation and Deployment}
As \textsc{Par4Sem} consists of different technologies, machine learning setups, resources, and configurations, we opted to provide a Docker-based installation and deployment options. While it is possible to fully install the tool on one’s own server, we also provide an API access for the whole backend services.  This allows users to quickly and easily install the frontend component and relay on our API service calls for the rest of the communications.
\section{Use-case -- Adaptive Text Simplification using Crowdsourcing}
\label{usecase}
	
An appropriate use case for adaptive paraphrasing is lexical text simplification. Lexical simplification aims to reduce the complexity of texts due to difficult words or phrases in the text \cite{Siddharthan2014}. We have used \textsc{Par4Sem} particularly for text simplification task with an emphasis of making texts accessible for language learners, children, and people with disabilities. 

\begin{figure*}
		\includegraphics[width=0.85\linewidth]{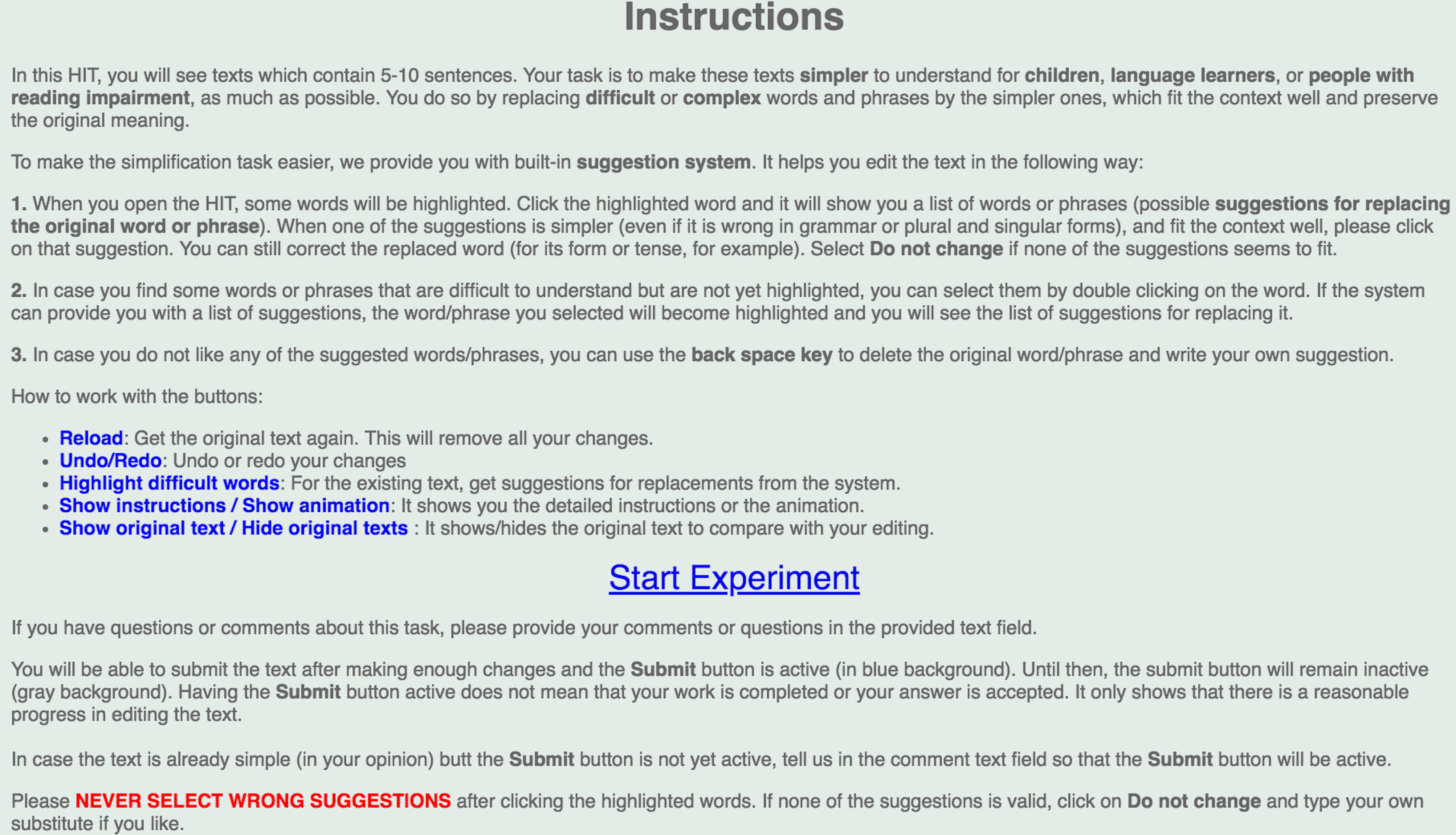}
		\caption{The instructions for the text simplification task using \textsc{Par4Sem}}
		\label{fig:par4semSimInst} 
\end{figure*}

We conducted the experiment by integrating the tool into the Amazon Mechanical Turk  (MTurk)\footnote{\url{https://www.mturk.com/}} crowdsourcing and employ workers to simplify texts using the integrated adaptive paraphrasing system. While \textsc{Par4Sem} is installed and run on our local server, we make use of the MTurk's external HIT  functionality to embed and conduct the text simplification experiment. Once workers have access to our embedded tool in the MTurk browser, they will be redirected to our local installation to complete the simplification task. Figure \ref{fig:par4semedit} shows the \textsc{Par4Sem} user interface to perform text simplification task by the workers while Figure \ref{fig:par4semSimInst} shows the instructions as they appeared inside the MTurk's browser. 
	\begin{figure}
		\includegraphics[width=1.0\linewidth, height=0.55\linewidth]{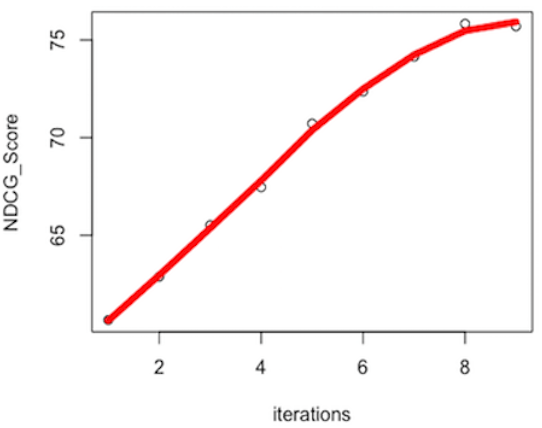}
		\caption{Learning curve showing the increase of NDCG@10 score over 9 iterations.}
		\label{fig:Learning_rate} 
\end{figure}

We asked workers to simplify the text for the target readers, by using the embedded paraphrasing system. Difficult words or phrases are automatically highlighted so that workers can click and see possible candidate suggestions. 
The experiment was conducted over 9 iterations, where the ranking model is updated using the training dataset (usage data) collected in the previous iterations. The first iteration does not use ranking model but candidates are presented using a default language-model-based ranking. 
In \cite{Yimam:2018:COLLING} we have shown that the adaptive paraphrasing system adopts very well to text simplification, improving the NDCG \cite{YiningWang2013} score from 60.66 to 75.70.
Figure \ref{fig:Learning_rate} shows the learning curve for the different iterations conducted in the experiment.

\section{Conclusion}
In this paper, we have described \textsc{Par4Sem}, a semantic writing aid tool based on an embedded adaptive paraphrasing system. Unlike most annotation tools, which are developed exclusively to collect training examples for machine learning applications, \textsc{Par4Sem} implements an adaptive paraphrasing system where training examples are obtained from usage data. 

 To the best of our knowledge, \textsc{Par4Sem} is the first of its kind where machine learning models are improved based on usage data and user feedback (correction of suggestions) for semantic applications. \textsc{Par4Sem} is used in a text simplification use-case. Evaluation of the system showed that the adaptive paraphrasing system for text simplification successfully adapted to the target task in a small number of iterations.

For future work, we would like to evaluate the system in an open task setting where users can paraphrase resp. simplify self-provided  texts, and explore how groups of similar users can be utilized to provide adaptations for their respective sub-goals. 

\bibliographystyle{acl_natbib_nourl}
\bibliography{emnlp2018}

\end{document}